# KBCNMUJAL@HASOC-Dravidian-CodeMix-FIRE2020:

# Using Machine Learning for Detection of Hate Speech and Offensive Code-Mixed Social Media text

Varsha Pathak[3], Manish Joshi[2], Prasad A. Joshi[3], Monica Mundada[4], Tanmay Joshi[5]

[1] Institute of Management and Research, Jalgaon, Affil- KBC North Maharashtra University, Jalgaon MS India

[2] School of Computer Sciences, KBC North Maharashtra University, Jalgaon, MS India

[3] JET's Z. B. College,, Dhule Affil- KBC North Maharashtra University, Jalgaon MS India

[4] Department of Dizitilization, Copenhagen Business School, Denmark

[5] BMCC college of commerce, Pune, India

**Abstract**

This paper describes the system submitted by our team KBCNMUJAL for Task 2 of the shared task "Hate Speech and Offensive Content Identification in Indo-European Languages (HASOC)" at FIRE 2020. The datasets of two Dravidian languages viz Malayalam and Tamil of size 4000 rows each, are shared by HASOC organizers. These datasets are used to train the machine using different machine learning algorithms, based on classification and regression models. The datasets consist of twitter messages with two class labels "offensive" and "not offensive". The machine is trained to classify such social media messages in these two categories. Appropriate n-gram feature sets are extracted to learn the specific characteristics of the Hate Speech text messages. These feature models are based on TF-IDF weights of n-gram. The referred work and respective experiments show that the features such as word, character and combined model of word and character n-grams could be used to identify the term patterns of offensive text contents.

As part of the HASOC shared task at FIRE 2020, the test data sets are made available by the HASOC track organisers. The best performing classification models developed for both languages, are applied on test datasets. The model which gives the highest accuracy result on training dataset for Malayalam language, was experimented to predict the labels of respective Test dataset. This system has obtained an F1 score of 0.77 and the model has received a HASOC rank of 2. Similarly the best performing model for Tamil language has obtained an F1 score of 0.87. It has received 3rd rank in the shared task participation of our team KBCNMUJAL. With the name of our team the system is named as *HASOC_kbcnmujal*.

## 1. Introduction

Social media has become a modern channel of public expression for the people irrespective of the socio economic boundaries. Due to the pandemic situations, even the common people have started looking at social media as a formal manner to remain connected with masses. Though such mediums are available majorly for constructive and creative expressions, these days it is found to be used for negative and offensive expressions. Some people take disadvantage of the language based social boundaries. They use absurd and hateful speech using their native languages to hurt other communities. Some people using hate speech or offensive language may be or may not deliberately but can hurt some popular person, specific community or even innocent people. If not detected in time, such messages could damage social health. Ignoring such messages could push certain unhealthy issues which could turn into disastrous events at a certain point. There are cases that such offensive comments have brought serious threats to the community disturbances. To identify such content is today's need and significant work is done in English language.

India is a multistate, multilingual nation. Each state has its own official spoken language and respective script. The transliterated Romanized text of the native language with English as the binding language is termed as Codemix Text. Many people use codemixed text to create their social media contents.

Most of the southern Indian languages have their origin in Dravidian language. Due to this the languages such as Malayalam of Kerala state, Tamil of Tamilnadu, Telugu of Andhra Pradesh are commonly called as Dravidian Languages. It's a challenge for the research community to trace and restrict such offensive content from the native codemixed text messages of Dravidian languages.

The "Hate Speech and Offensive Content Identification in Indo-European Languages (HASOC)" at FIRE 2019 [1] is the first such initiatives as a shared task on offensive language. The HASOC track has further introduced the shared task on Dravidian Code-Mixed text in FIRE 2020 [10], including Malayalam and Tamil. The goal of Task 2 is to classify the tweets into offensive or not-offensive categories. This paper presents the result of our model for Task 2.

## 2. Related Work

Many researchers have published their work on automated detection of hate speech and offensive content. Malmasi and Zampieri [2] used a linear Support Vector Classifier on word skip-grams, brown cluster and Surface n-grams. Arup Baruah et al. [3] used Support Vector Machine, BiLSTM and Neural Network models on TF-IDF features of character and word n-grams, Embeddings from language models, Glove and fastText embeddings. Anita Saroj et al. [8] used traditional machine learning classifiers: Support Vector Machine, XGBOOST. Nemanja Djuric et al. [4] used paragraph2vec and Continuous Bag of Words to approach the neural language model.

Recently a few studies show work on detecting hate speech other than English language. The system developed by Mubarak et al. [5] uses SeedWords, word unigrams, word bigrams approach to detect abusive language in Arabic social media. Su et al. [6] develops a model to detect and rephrase profanity in Chinese. Anita and Sukomal [7] apply Machine learning classifiers on TF-IDF features on comments in Hindi and English languages. Bharathi et al. [9] used many machine learning classifiers to determine the sentiments from Malayalam-English code-mixed data.

## 3. Hate Speech Datasets

The datasets were available in Manglish (Malayalam English) and Tanglish (Tamil English) [11] languages. Training and test datasets were released for Task2. Both the training datasets have 3 columns: first is ID, second is tweets or YouTube comments and the last column is of labels indicating either offensive (OFF) or non-offensive (NOT) tweet. Test datasets of both languages were released later that contained IDs and tweets text. The Label column was missing. For both languages, the training set consisted of 4000 tweets and Test set had approximately 1000 tweets. Table 1 presents the statistical data about this Training and Test Data Set for both Malayalam and Tamil languages. The details of how these datasets are constructed and the overview of the shared HASOC tasks is available in [12].

Table 1. Training Set and Test Set Statistics for Malayalam

| Indian Language | Dataset | Type of Tweet | % | Total |
|---|---|---|---|---|
| Malayalam | Training | Not Offensive | 2047 (51.18%) | 4000 |
| | Training | Offensive | 1953 (48.82%) | |
| | Test | Not Offensive | Not Known | |
| | Test | Offensive | Not Known | 1000 |
| Tamil | Training | Not Offensive | 2020 (50.5%) | 4000 |
| | Training | Offensive | 1980 (49.5%) | |
| | Test | Not Offensive | Not Known | |
| | Test | Offensive | Not Known | 940 |

## 4. Methodology

In this work supervised machine learning is used by experimenting on various Classifiers. The labeled datasets were preprocessed for removal of noisy elements from its contents. Appropriate features are extracted to enable the machine to learn offensive term patterns.

Finally the performances of different Classifiers and feature models are compared using standard measures to choose the best performing model. We call this system with our team name as **HASOC_kbcnmujal** system.

### 4.1 Data Preprocessing

In general, Social media text does not obey grammatical rules and are written in more than one language [12]. Both training datasets experimented primarily by removing emojis and Unicode symbols because they are not helping in enhancing the performance of our model. We have also tested that the English stop words are not improving the result, so removed them. Digits, special characters [@,#,%,$,^,(,)], hyphens, extra white spaces and null values are also removed. In addition to above we sorted the Tamil dataset as per labels column, because they are in serial order and also removed @USER, @RT(retweet) and #TAG.

All these preprocessing has removed most of the noisy and unrequired elements from the tweeter text. As a result, the datasets for training are now in pure Manglish and Tanglish language tweets.

### 4.2 Features Extraction

For feature extraction we applied two major methods viz. TF-IDF and Custom Word Embedding methods. The details of these methods are as follows.

### 4.2.1 Using TF-IDF

The Term Frequency (TF) and Inverse Document Term Frequency (IDF) are the two important measures that reflect the specificity and relevance of terms with the information carried by the documents. We have used TF-IDF weights for the n-gram features extracted from the tweets in the datasets.

For Malayalam Language our model uses word n-grams of order (1, 2). This has extracted 38536 features. Similarly from character n-grams of order (1, 5) extracted 81191 features. The combined word n-grams (1, 2) and character n-grams (1, 5) has extracted 119727 features. All these three feature extraction methods were experimented to choose the best performing feature model.

In case of the Tamil language all the above three feature models were applied. The word n-grams of order (1, 4) has extracted 117173 features. The character n-grams of order (1, 7) has extracted 325902 features and the combined word n-grams (1, 4) and character n-grams (1, 7) has extracted 443075 features.

These n-grams are useful to capture the small and localized syntactic patterns within text in flexible language. Table 2 and 3 respectively present a sample of these features in Malayalam and Tamil languages text occurring in the corresponding datasets.

Table 2. Features using TF-IDF

| Malayalam Language | |
|---|---|
| Char Features | Word Features |
| acho', ' acr', ' acra', ' act', ' act ', ' acti', ' actn', ' acto', ' actu', ' ad', ' ada', ' ada ', ' adal', ' adan', ' adap', ' add', ' adde', ' | 'aa', 'aa aala', 'aa accountinte', 'aa adikkan', 'aa adisthanathil', 'aa al', 'aa amma', 'aa bagyavan', 'aa basheerine', 'aa bedinadiyale', 'aa beef', 'aa casaba', 'aa |

| addr' ' ade' ' adec', ' adee', ' ades', ' adh', ' adh ', ' adha', ' adhe',... | chekkanitta', 'aa cinema', 'aa collegl', 'aa coment', 'aa duplicate', 'aa ellam', 'aa film', 'aa indarw',... |
|---|---|

Table 3. Features using TF-IDF

| Tamil Language ||
|---|---|
| Char Features | Word Features |
| ' aa', ' aa ', ' aa a', ' aa at',' aa ath', ' aa d', ' aa da', ' aa da ', ' aa g', ' aa ga', ' aa gay', ' aa i', ' aa ir', ' aa iru', ' aa k', ' aa ka', ' aa kaa', ' aa ku', ' aa ku ',... | 'aa iruku athanala', 'aa iruku athanala na', 'aa irukum', 'aa irukum ippo', 'aa irukum ippo karaname', 'aa kaatuthu', 'aa kaatuthu konjam', 'aa kaatuthu konjam decent', 'aa ku', 'aa ku romba', 'aa ku romba kastam', 'aa kuppa', 'aa kuppa thottiyil', 'aa lossu'.... |

### 4.2.2 Using Custom Word Embedding

We extracted 15430 & 15292 unique words from Malayalam and Tamil respectively, some of the unique words are listed in Table 4. Then we found the length of longest sentence from both languages and we got longest length as 65 for Malayalam and for Tamil longest length is 64. In the next step we made all sentences of equal size i.e. equal to the longest sentence of that language. To achieve this we append zeros at the end of each sentence using pad_sequences method from *Keras[2]*. Thus we have created custom word embedding for both languages.

Table 4. Unique words

| Malayalam Language | Tamil Language |
|---|---|
| Unique words | Unique words |
| 'monjum', 'anthu', 'da', 'mathi','aarudeyum', 'manju', 'vandi','analla','Illee', 'kandotte','pottanum', 'demand','ondakkalle', 'busy','kettapol', 'cheyyan','nenthrakaya', 'kottayam', 'undakki'.... | 'ootunathu', 'korona', 'netha','Pannuvaen', 'kondadunom','Ad', 'teriyala', 'arasiyal','potrukkanga', 'yudham','neeyalam', 'aiten', 'kandara', 'epdiii', 'Iyaooo', 'pathathum', 'suriya', 'okuarthuku', 'next','kaattave', 'peasuriya',... |

## 4.3 Experimental work

Model uses a supervised classification approach. The training data are labeled with 2 classes OFF or NOT for either "offensive" or "not-offensive" tweets respectively.

### 4.3.1 Classifier Models

We used various classifiers like: Support Vector Classifier (SVC), Multinomial Naive Bayes (MNB), Logistic Regression (LR), AdaBoost, Decision Tree Classifier (DTC) and Random Forest Classifier (RFC). Using the above mentioned extracted features we trained our machine learning classifiers. We have investigated the hyper parameters of each classifier and found the best parameter value for both languages. For evaluation we used *sklearn*[3]. Among the above classifiers who performed extremely well are described below.

---

[2] https://keras.io/

For both the languages using custom word embedding, we trained our text classification neural network model [15].

Support Vector Classifier (SVC)-
We used Linear SVC. SVC fits the data, and returns a best fit hyperplane that divides the data points. It scales large number of samples and has more flexibility in the choice of penalties and loss function [13]. Basically SVC is a binary classifier. In case of multi class dataset it uses One vs. Rest (OvR) strategy. For Malayalam, SVC was trained using the TF-IDF features of word n-grams (1, 2), character n-grams (1, 5) and combined word n-grams (1, 2) and character n-grams (1, 5). For Tamil, we trained SVC using the TF-IDF features of word n-grams (1, 4), character n-grams (1, 7) and combine word n-grams (1, 4) and character n-grams (1, 7). For both languages we tuned the parameters: kernel as 'linear', 'rbf' and gamma as 'auto', 'scale'. The L2 regularization was used and the hyperparameter C was also investigated for extracted features. For both the languages we found that the character n-grams features increased the accuracy of the classifier as compared to word n-gram and word plus character n-grams features models.

Multinomial Naive Bayes (MNB):
Multinomial Naive Bayes is a probabilistic model and specialized version of Naive Bayes. Simple Naive Bayes model a document as the presence and absence of particular words, whereas Multinomial Naive Bayes explicitly models the word counts and adjusts the underlying calculations to deal with in [14]. It works very well on small amounts of training data and gets trained relatively fast compared to other models. The MNB was trained using the same TF-IDF features for both language datasets as mentioned above. We have tested the hyper-parameter alpha with different values. Combining the word n-gram and character n-gram feature gives best accuracy in case of both Tamil and Malayalam datasets,

Ensemble Approach:
We used a hard voting approach. Hard voting sums the predicates for each class label from multiple models and predicts the class label with maximum votes. We combined the predictions of our top three models: SVM, MNB & LR. Combining word n-gram and character n-gram feature gives best accuracy for datasets of both languages.

Logistic Regression (LR):
Logistic Regression is a statistical model. It transforms its output into a probability value which can be mapped to two or more discrete classes. Logistic regression is the regression analysis used to conduct when the dependent variable is binary.
We trained LR in the same way as that of SVC and MNB. L2 regularization was used and the hyper-parameter C was set to default i.e. 1.0. Combined word and character n-gram features using TF-IDF weights give higher accuracy for both languages.

Random Forest Classifiers (RFC):
It is an ensemble approach combining multiple decision trees and producing them randomly without defining the rules [9]. We keep all the hyper-parameters to default only we tested n-estimators to different values. We found that character n-gram feature and combine word and character n-gram features has near about the same accuracies for both datasets.

Neural Network for Text Classification:
Simple text classification Neural Network model is created using python's *Keras Library*. *Keras* is one of the most famous and commonly used deep learning libraries. It can be used to learn custom words embedding or used

---
[3] http://scikit-learn.org/

to load pre-trained word embedding. We used Keras Sequential model and add embedding layer as a first layer. *Keras* embedding layer takes 3 parameters as arguments as shown below:

*keras.layers.Embedding(size_of_vocabulary,number_of_word_dimensions,length_of_longest_sentence)*

Here, for both dataset *size_of_vocabulary* is nothing but the number of unique words that we have already extracted i.e. 15430(Malayalam) & 15292(Tamil). At the time of training the model, we respectively rounded these values as 15450 & 15300. The Second parameter *number_of_word_dimensions* represents each word as a 200 dimensional vector. And the third parameter *length_of_longest_sentence* is length of longest sentence from dataset i.e. 65(Malayalam) and 64(Tamil).

After creating the model for both the languages, at the embedding layer we got 30,90,000 trainable parameters for Malayalam and 30,60,000 for Tamil. The output of the embedding layer is a sentence with 65 words for Malayalam and for Tamil it is 64 words; where each word is represented by a 200 dimensional vector. Then we flattened the embedding layer and we got 13000 dimensional vectors for Malayalam and 12800 for Tamil. How trainable parameters and dimensional vectors are achieved is shown below in Table 5.

Table 5. Trainable Parameters and Dimensional vectors for Malayalam and Tamil

| For Malayalam | At Embedding layer | 15450 Unique words | X | 200 dimensional vector | = | 30,90,000 Trainable Parameters |
|---|---|---|---|---|---|---|
| | Output of Embedding Layer | 65 Words | X | 200 dimensional vector | = | 13,000 Dimensional Vectors |
| For Tamil | At Embedding layer | 15300 Unique words | X | 200 dimensional vector | = | 30,60,000 Trainable Parameters |
| | Output of Embedding Layer | 64 Words | X | 200 dimensional vector | = | 12,800 Dimensional Vectors |

The second layer i.e. dense layer has 1 neuron. Since ours is a binary classification problem, we use the sigmoid function as the loss function at the dense layer. The Adam optimizer and the binary cross-entropy loss function were used for training.

## 5. Results

For evaluating the performance of our model, we have used 5-fold cross-validation. 30% of the dataset was used as test dataset and 70% was used for training the model and measuring the results in terms of accuracy. For both the languages we used the same strategy.

The performance of classifier's accuracy versus extracted features for Malayalam and Tamil Language is shown in Figure1 and Figure 2 respectively.

From Figure 1 & 2, we state that the accuracy performance of MNB, LR and Ensemble is in increasing order for word, character and combined word and character n-gram features for both the languages. In case of SVC for both languages, the character n-gram increases the accuracy as compared to the other two features. Overall RFC has performed very low.

In Table 6 and Table 7 we have shown accuracy reports of classifiers for offensive and not offensive contents for both Malayalam and Tamil language respectively.

From Table 6 we can see that the Malayalam language MNB has an F1 score of 78 for "not offensive" tweets, and for "offensive" tweets has the highest F1 score of 74. Hence MNB stood out as the best classifier. The second best F1 score of 72 for "offensive" tweets is for SVC, LR and Ensemble. Among the classifiers performance of SVC & LR are same for "offensive" and "not offensive" tweets. RFC performs very low in predicting both the categories. Simple NN Model scored 0.52.

For Tamil language we can see in Table 7 SVC has the highest F1 score of 87 and 86 for "not offensive", and "offensive" tweets respectively. The Performances of MNB, LR and Ensemble are identical. RFC has also performed well for Tamil language. Simple NN Model did not make any improvement for Tamil and scored 0.53.

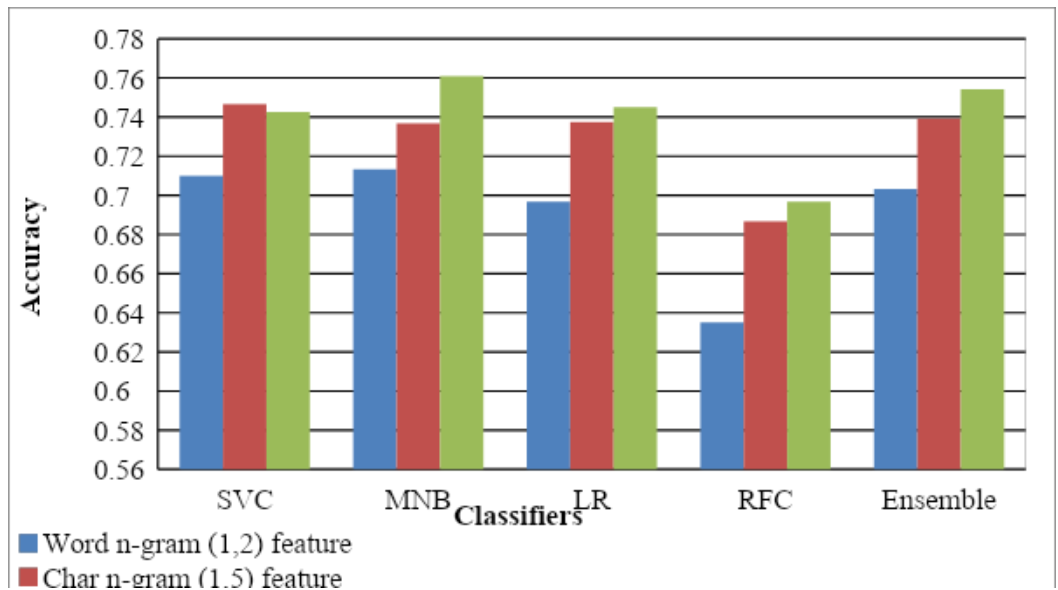

**Figure 1:** Performance of classifiers as per features in Malayalam Language

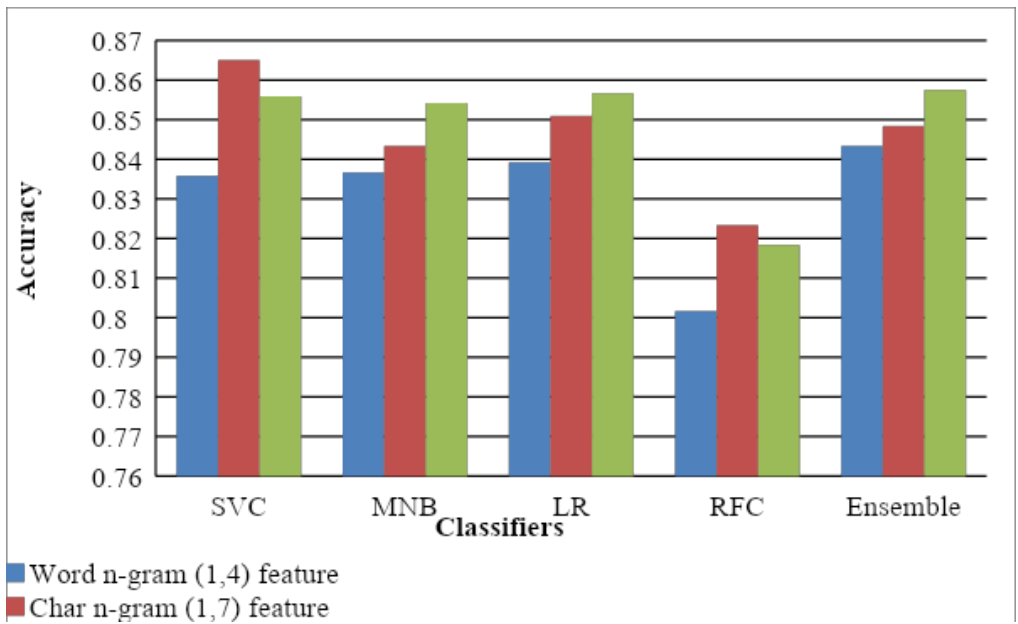

**Figure 2:** Performance of classifiers as per features in Tamil Language.

**Table 6: Result of models for Malayalam language**

| Classifier | Features | Accuracy | Precision | Recall | F1 |
|---|---|---|---|---|---|
| SVC | Word n-gram (1,2) | 0.71 | 0.72 | 0.64 | 0.68 |
| SVC | Char n-gram (1,5) | 0.75 | 0.76 | 0.69 | 0.72 |
| SVC | Combine Char & Word n-gram | 0.74 | 0.75 | 0.69 | 0.72 |
| MNB | Word n-gram (1,2) | 0.71 | 0.73 | 0.63 | 0.68 |
| MNB | Char n-gram (1,5) | 0.74 | 0.74 | 0.70 | 0.72 |
| **MNB** | **Combine Char & Word n-gram** | **0.7608** | **0.77** | **0.72** | **0.74** |
| LR | Word n-gram (1,2) | 0.70 | 0.72 | 0.60 | 0.65 |
| LR | Char n-gram (1,5) | 0.74 | 0.75 | 0.68 | 0.71 |
| LR | Combine Char & Word n-gram | 0.75 | 0.76 | 0.68 | 0.72 |
| RFC | Word n-gram (1,2) | 0.66 | 0.62 | 0.61 | 0.62 |
| RFC | Char n-gram (1,5) | 0.69 | 0.71 | 0.59 | 0.64 |
| RFC | Combine Char & Word n-gram | 0.70 | 0.73 | 0.59 | 0.65 |
| Ensemble | Word n-gram (1,2) | 0.70 | 0.75 | 0.57 | 0.65 |
| Ensemble | Char n-gram (1,5) | 0.74 | 0.78 | 0.64 | 0.70 |
| Ensemble | Combine Char & Word n-gram | 0.75 | 0.80 | 0.65 | 0.72 |
| NN Model | Custom Word Embedding | 0.52 | 0.49 | 0.42 | 0.45 |

**Table 7: Result of models for Tamil language**

| Classifier | Features | Accuracy | Precision | Recall | F1 |
|---|---|---|---|---|---|
| SVC | Word n-gram (1,4) | 0.84 | 0.83 | 0.84 | 0.83 |
| **SVC** | **Char n-gram (1,7)** | **0.87** | **0.86** | **0.86** | **0.86** |
| SVC | Combine Char & Word n-gram | 0.86 | 0.86 | 0.85 | 0.85 |
| MNB | Word n-gram (1,4) | 0.84 | 0.83 | 0.84 | 0.83 |
| MNB | Char n-gram (1,7) | 0.84 | 0.82 | 0.87 | 0.84 |
| MNB | Combine Char & Word n-gram | 0.85 | 0.83 | 0.87 | 0.85 |
| LR | Word n-gram (1,4) | 0.84 | 0.84 | 0.83 | 0.83 |
| LR | Char n-gram (1,7) | 0.85 | 0.86 | 0.82 | 0.84 |
| LR | Combine Char & Word n-gram | 0.86 | 0.87 | 0.84 | 0.85 |
| RFC | Word n-gram (1,4) | 0.80 | 0.80 | 0.78 | 0.79 |
| RFC | Char n-gram (1,7) | 0.82 | 0.86 | 0.77 | 0.81 |
| RFC | Combine Char & Word n-gram | 0.81 | 0.85 | 0.76 | 0.80 |
| Ensemble | Word n-gram (1,4) | 0.85 | 0.85 | 0.82 | 0.84 |
| Ensemble | Char n-gram (1,7) | 0.85 | 0.86 | 0.82 | 0.84 |
| Ensemble | Combine Char & Word n-gram | 0.86 | 0.87 | 0.83 | 0.85 |
| NN Model | Custom Word Embedding | 0.53 | 0.52 | 0.51 | 0.52 |

From Table 8 for Malayalam language, we can straightforwardly see that using the combined char & word n-grams feature MNB predicts "not offensive" & "offensive" tweets in the best way. The SVC, LR and Ensemble have high predictions in "not offensive" but lack in predicting "offensive" tweets. The noticeable thing is combining char & word n-grams features have increased the performance for all classifiers except SVC in "not offensive" and for RFC in "offensive" category.

Table 9 is for Tamil language. For predicting "not offensive" category LR and Ensemble performance is high, but unable to score for "offensive" category. Performance of MNB, for predicting "offensive" category tweets is high for each feature. As we can see, the results of Char n-grams and combining char and word n-gram features for SVC are marginally the same, but Char n-gram got best predictions.

**Table 8: Confusion matrix for "Not Offensive" & "Offensive" tweets in Malayalam language**

| | SVC Word n-grams | | SVC Char n-grams | | SVC Char & Word n-grams | | MNB Word n-grams | | MNB Char n-grams | | **MNB Char & Word n-grams** | | LR Word n-grams | | LR Char n-grams | | LR Char & Word n-grams | |
|---|---|---|---|---|---|---|---|---|---|---|---|---|---|---|---|---|---|---|
| NOT | 484 | 140 | 500 | 124 | 493 | 131 | 491 | 133 | 479 | 145 | **499** | **125** | 492 | 132 | 496 | 128 | 501 | 123 |
| OFF | 208 | 368 | 180 | 396 | 178 | 398 | 211 | 365 | 171 | 405 | **162** | **414** | 232 | 344 | 187 | 389 | 183 | 393 |

| | RFC Word n-grams | | RFC Char n-grams | | RFC Char & Word n-grams | | Ensemble Word n-grams | | Ensemble Char n-grams | | Ensemble Char & Word n-grams | | NN Model | |
|---|---|---|---|---|---|---|---|---|---|---|---|---|---|---|
| NOT | 409 | 215 | 484 | 140 | 497 | 127 | 517 | 107 | 518 | 106 | 530 | 94 | 377 | 247 |
| OFF | 223 | 353 | 236 | 340 | 237 | 339 | 249 | 327 | 207 | 369 | 201 | 375 | 335 | 241 |

**Table 9: Confusion matrix for "Not Offensive" & "Offensive" tweets in Tamil language**

|  | SVC Word n-grams | | **SVC Char n-grams** | | SVC Char & Word n-gram | | MNB Word n-gram | | MNB Char n-gram | | MNB Char & Word n-gram | | LR Word n-grams | | LR Char n-grams | | LR Char & Word n-gram | |
|---|---|---|---|---|---|---|---|---|---|---|---|---|---|---|---|---|---|---|
| NOT | 514 | 103 | **539** | **78** | 534 | 83 | 516 | 101 | 504 | 113 | 516 | 101 | 532 | 94 | 541 | 76 | 541 | 76 |
| OFF | 94 | 489 | **84** | **499** | 90 | 493 | 95 | 488 | 75 | 508 | 74 | 509 | 99 | 484 | 103 | 480 | 96 | 487 |

|  | RFC Word n-grams | | RFC Char n-grams | | RFC Char & Word n-gram | | Ensemble Word n-grams | | Ensemble Char n-grams | | Ensemble Char & Word n-gram | | NN Model | |
|---|---|---|---|---|---|---|---|---|---|---|---|---|---|---|
| NOT | 505 | 112 | 542 | 75 | 539 | 78 | 536 | 81 | 542 | 75 | 544 | 73 | 339 | 278 |
| OFF | 126 | 457 | 137 | 446 | 140 | 443 | 107 | 476 | 107 | 476 | 98 | 485 | 284 | 299 |

Later we combined the Malayalam and Tamil datasets and applied the same model and checked how the system performs. While combining two approaches, first we put the datasets one after another. In second approach sorted the tweets column in ascending to descending order, so that tweets are not serially. We extracted the features by applying the same method of TF-IDF, but we changed the n-gram range. So TF-IDF weights of word n-gram (1, 6) and character n-gram (1, 8) were applied. We only applied MNB, SVC, LR and Ensemble hard voting approaches.

For the first approach of combining the dataset i.e. placing one after another, MNB, SVC, LR and Ensemble got the highest accuracy score of 0.80 for the combined word and character n-gram features. On the other for the second approach, where we sorted the dataset, the accuracy of each classifier for combined word and character n-gram features drops by 1% except SVC keeps the same accuracy of 0.80. The compared performance of classifiers is shown in Figure 3.

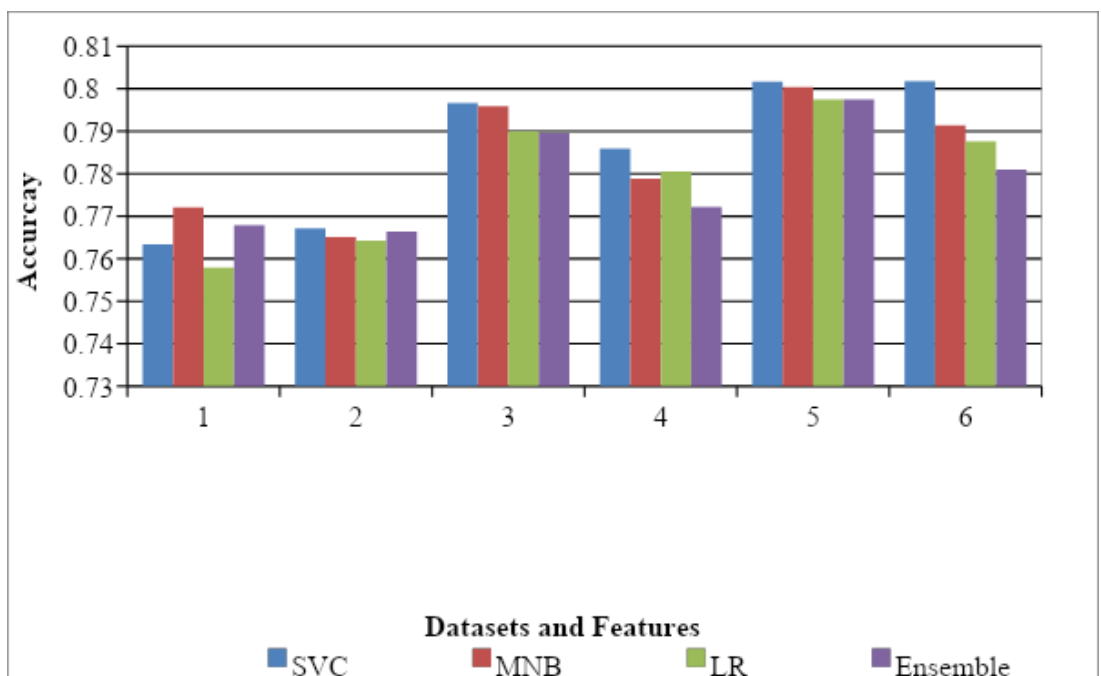
**Figure 3:** Performance of classifiers on combined Datasets

## 6. Conclusion

This paper presents the experimental work and respective results of the task to detect offensive content in code-mixed dataset of Dravidian languages. We used different features: word n-gram, character n-gram, combined word, character n-grams and custom word embedding. Using the TF-IDF weights of word and character n-gram features we trained our machine learning classifiers. Custom word embedding was used to train a simple neural network model. Our model for Malayalam language got the official rank of 2$^{nd}$ and obtained an F1 score of 0.77. For Tamil language Model we got the official rank of 3rd and F1 score of 0.87.

This work will be further extended to develop a system that could learn offensive terms from the text contents or even from speech irrespective of the language. We are interested in revealing hidden negative messages from the social media comments which may sound superficially positive. Such contents that can damage the social and communal health shall be cured at the right time.